\pdfoutput=1

\documentclass[11pt,a4paper]{article}
\usepackage[hyperref]{AACL-IJCNLP2020}
\usepackage{times}
\usepackage{latexsym}

\usepackage{microtype}

\usepackage{todonotes}
\usepackage{dirtytalk}
\usepackage[ruled, linesnumbered]{algorithm2e}
\usepackage{todonotes}
\aclfinalcopy 

\title{Data Augmentation Using Pre-trained Transformer Models}

\author{Varun Kumar  \\
  Alexa AI \\
  \texttt{kuvrun@amazon.com} \\\And
  Ashutosh Choudhary
 \\
  Alexa AI \\
  \texttt{ashutoch@amazon.com}
  \\\And
  Eunah Cho \\ 
  Alexa AI \\ 
  \texttt{eunahch@amazon.com}
  \\}

\date{}

\begin{document}
\maketitle

\begin{abstract}
Language model based pre-trained models such as BERT have provided significant gains across different NLP tasks.  In this paper, we study different types of transformer based pre-trained models such as auto-regressive models (GPT-2), auto-encoder models (BERT), and seq2seq models (BART) for conditional data augmentation. We show that prepending the class labels to text sequences provides a simple yet effective way to condition the pre-trained models for data augmentation. Additionally, on three classification benchmarks, pre-trained Seq2Seq model outperforms other data augmentation methods in a low-resource setting. Further, we explore how different data augmentation methods using pre-trained model differ in-terms of data diversity, and how well such methods preserve the class-label information.   
\end{abstract}
\section{Introduction}

Data augmentation (DA) is a widely used technique to increase the size of the training data. Increasing training data size is often essential to reduce overfitting and enhance the robustness of machine learning models in low-data regime tasks.  

In natural language processing (NLP), several word replacement based methods have been explored for data augmentation. In particular, ~\citet{wei2019eda} showed that simple word replacement using knowledge bases like WordNet~\cite{miller1998wordnet} improves classification performance. Further, ~\citet{kobayashi2018contextual} utilized language models (LM) to augment training data. However, such methods struggle with preserving class labels. For example, non-conditional DA for an input sentence of sentiment classification task \textit{``a small impact with a big movie"} leads to \textit{``a small movie with a big impact"}. Using such augmented data for training, with the original input sentence's label (i.e. negative sentiment in this example) would negatively impact the performance of the resulting model. 

To alleviate this issue, ~\citet{wu2019conditional} proposed conditional BERT (CBERT) model which extends BERT ~\cite{devlin2018bert} masked language modeling (MLM) task, by considering class labels to predict the masked tokens. Since their method relies on modifying BERT model's segment embedding, it cannot be generalized to other pre-trained LMs without segment embeddings. 

Similarly, ~\citet{anaby2019not} used GPT2~\cite{radford2019language} for DA where examples are generated for a given class by providing class as input to a fine-tuned model. In their work, GPT2 is used to generate $10$ times the number of examples required for augmentation and then the generated sentences are selected based on the model confidence score. As data selection is applied only to GPT2 but not to the other models, the augmentation methods can not be fairly compared. Due to such discrepancies, it is not straightforward to comprehend how the generated data using different pre-trained models varies from each other and their impact on downstream model performance. 

This paper proposes a unified approach to use any pre-trained transformer \cite{vaswani2017attention} based models for data augmentation. In particular, we explore three different pre-trained model types for DA, including 1) an autoencoder (AE) LM: BERT, 2) an auto-regressive (AR) LM: GPT2, and 3) a pre-trained seq2seq model: BART~\cite{lewis2019bart}. We apply the data generation for three different NLP tasks: sentiment classification, intent classification, and question classification. 

In order to understand the significance of DA, we simulate a low-resource data scenario, where we utilize only 10 training examples per class in a classification task. Section \ref{datasets} provides details of the task and corpora. 

We show that all three types of pre-trained models can be effectively used for DA, and using the generated data leads to improvement in classification performance in the low-data regime setting. Among three types of methods, pre-trained seq2seq model provides the best performance. Our code is available at \footnote{\url{https://github.com/varinf/TransformersDataAugmentation}}. 

Our contribution is three-fold: (1) implementation of a seq2seq pre-trained model based data augmentation, (2) experimental comparison of different data augmentation methods using conditional pre-trained model, (3) a unified data augmentation approach with practical guidelines for using different types of pre-trained models.

\section{DA using Pre-trained Models}

LM pre-training has been studied extensively ~\cite{radford2018improving, devlin2018bert, liu2019roberta}. During pre-training, such models are either trained in an AE setting or in an AR setting. In the AE setting, certain tokens are masked in the sentence and the model predicts those tokens. In an AR setting, the model predicts the next word given a context. Recently, pre-training for seq2seq model has been explored where a seq2seq model is trained for denoising AE tasks ~\cite{lewis2019bart, raffel2019exploring}. Here, we explore how these models can be used for DA to potentially improve text classification accuracy.

\begin{algorithm}[th]
\SetAlgoLined
\SetKwInOut{Input}{Input}
 \Input{Training Dataset $ \scriptstyle D_{train}$ \\ Pretrained model $ \scriptstyle G \in \{AE, AR, Seq2Seq\}$ }
 Fine-tune $\scriptstyle G$ using $ \scriptstyle D_{train}$ to obtain $\scriptstyle G_{tuned}$ \\
$ \scriptstyle D_{synthetic} \gets \{\}$ \\
\ForEach{ $ \scriptstyle {\{x_i, y_i\}} \in D_{train}$}{
Synthesize $s$ examples $\scriptstyle{\{\hat{x_i}, \hat{y_i}\}}_p^1$ using $\scriptstyle G_{tuned}$ \\
$ \scriptstyle D_{synthetic} \gets D_{synthetic} \cup {\{\hat{x_i}, \hat{y_i}\}}_p^1 $
}
 \caption{\label{da_algo} Data Augmentation approach} 
\end{algorithm}

\textbf{DA Problem formulation}: Given a training dataset $D_{train} = {\{x_i, y_i\}}_n^1$, where $x_i ={\{w_j\}}_m^1$ is a sequence of $m$ words, $y_i$ is the associated label, and a pre-trained model $G$, we want to generate a dataset of $D_{synthetic}$. Algorithm \ref{da_algo} describes the data generation process. For all augmentation methods, we generate $s=1$ synthetic example for every example in $D_{train}$. Thus, the augmented data is same size as the size of the original data. 

\subsection{Conditional DA using Pre-trained LM}
\label{approach_bert}
For conditional DA, a model $G$ incorporates label information during fine-tuning for data generation.
~\citet{wu2019conditional} proposed CBERT model where they utilized BERT's segment embeddings to condition model on the labels. Similarly, models can be conditioned on labels by prepending labels $y_i$ to $x_i$~\cite{keskar2019ctrl,johnson2017google}. 

Due to segment embedding reuse, CBERT conditioning is very specific to BERT architecture thus cannot be applied directly to other pre-trained LMs. Thus, we compare two generic ways to condition a pre-trained model on class label: 
\begin{itemize} 
\item \texttt{prepend} : prepending label $y_i$ to each sequence $x_i$ in the training data without adding $y_i$ to model vocabulary 
\item \texttt{expand} : prepending label $y_i$ to each sequence $x_i$ in the training data and adding $y_i$ to model vocabulary. 
\end{itemize} 
Note that in \texttt{prepend}, the model may split $y_i$ into multiple subword units \cite{sennrich2015neural, kudo2018sentencepiece}, \texttt{expand} treats a label as a single token. 

Here, we discuss the fine-tuning and the data generation process for both AE and AR LMs. For transformer based LM implementation, we use Pytorch based transformer package \cite{Wolf2019HuggingFacesTS}. For all pre-trained models, during fine-tuning, we further train the learnable parameters of $G$ using its default task and loss function.

\subsubsection{Fine-tuning and generation using AE LMs}
We choose BERT as a representative of AE models. For fine-tuning, we use the default masking parameters and MLM objective which randomly masks some of the tokens from the raw sequence, and the objective is to predict the original token of the masked words using the context. Both BERT\textsubscript{prepend} and BERT\textsubscript{expand} models are fine-tuned using the same objective.  

\subsubsection{Fine-tuning and generation using AR LMs}
For AR LM experiments, we choose GPT2 as a generator model and follow the method proposed by ~\citet{anaby2019not} to fine-tune and generate data. For fine-tuning GPT2, we create a training dataset by concatenating all sequences in $D_{train}$ as follows: 
$y_1 SEP x_1 EOS y_2 ... y_n SEP x_n EOS$. $SEP$ denotes a separation token between label and sentence, and $EOS$ denotes the end of a sentence. 

For generating data, we provide $y_i SEP$ as a prompt to $G$, and we keep generating until the model produces $EOS$ token. We use GPT2 to refer to this model. We found that such generation struggles in preserving the label information, and a simple way to improve the generated data label quality is to provide an additional context to $G$. Formally, we provide $y_i SEP w_1.. w_k$ as prompt where $w_1.. w_k$ are the first $k$ words of a sequence $x_i$. In this work, we use $k=3$. We call this method GPT2\textsubscript{context}. 

\subsection{Conditional DA using Pre-trained Seq2Seq model}
Like pre-trained LM models, pre-training seq2seq models such as T5~\cite{raffel2019exploring} and BART~\cite{lewis2019bart} have shown to improve performance across NLP tasks. 
For DA experiments, we choose BART as a pre-trained seq2seq model representative for its relatively lower computational cost. 

\subsubsection{Fine-tuning and generation using Seq2Seq BART}
Similar to pre-trained LMs, we condition BART by prepending class labels to all examples of a given class. While BART can be trained with different denoising tasks including insertion, deletion, and masking, preliminary experiments showed that masking performs better than others. Note that masking can be applied at either word or subword level. We explored both ways of masking and found subword masking to be consistently inferior to the word level masking. Finally, we applied word level masking in two ways: 
\begin{itemize} 
\item BART\textsubscript{word} : Replace a word $w_i$ with a mask token $<mask>$ 
\item BART\textsubscript{span}: Replace a continuous chunk of $k$ words $w_i, w_{i+1}..w_{i+k}$ with a single mask token $<mask>$. 
\end{itemize} 
Masking was applied to 40\% of the words. We fine-tune BART with a denoising objective where the goal is to decode the original sequence given a masked sequence.  


\subsection{Pre-trained Model Implementation}
\label{appendixa3} 
\subsubsection{BERT based DA models}
For AutoEncoder (AE) experiments, we use \say{bert-base-uncased} model with the default parameters provided in huggingface's transformer package. In \texttt{prepend} setting we train model for 10 epochs and select the best performing model on dev data partition keeping initial learning rate at $4e^{-5}$. For \texttt{expand} setting, training requires 150 epochs to converge. Moreover, a higher learning rate of $1.5e^{-4}$ was used for all three datasets. The initial learning rate was adjusted for faster convergence. This is needed for \texttt{expand} setting as embeddings for labels are randomly initialized.

\subsubsection{GPT2 model implementation}
For GPT2 experiments, we use GPT2-Small model provides in huggingface's transformer package. We use default training parameters to fine-tune the GPT2 model. For all experiments, we use $SEP$ as a separate token and $<\mid endoftext\mid>$ as EOS token. For text generation, we use the default nucleus sampling~\cite{holtzman2019curious} parameters including $top\_k=0$, and $top\_p = 0.9$.  

\subsubsection{BART model implementation}
For BART model implementation, we use fairseq toolkit~\cite{ott2019fairseq} implementation of BART. Additionally, we used bart\_large model weights\footnote{\url{https://dl.fbaipublicfiles.com/fairseq/models/bart.large.tar.gz}}. 

Since BART model already contains $<mask>$ token, we use it to replace mask words. For BART model fine-tuning, we use denoising reconstruction task where 40\% words are masked and the goal of the decoder is to reconstruct the original sequence. Note that the label $y_i$ is prepended to each sequence $x_i$, and the decoder also produces the label $y_i$ as any other token in $x_i$. We use fairseq's \textit{label\_smoothed\_cross\_entropy} criterion with a \textit{label-smoothing} of $0.1$. We use $1e^{-5}$ as learning rate. For generation, beam search with a beam size of $5$ is used. 

\subsection{Base classifier implementation}
\label{base_classifier}
For the text classifier, we use \say{bert-base-uncased} model. The BERT model has $12$ layers, $768$ hidden states, and $12$ heads. We use the pooled representation of the hidden state of the first special token ([CLS]) as the sentence representation. A dropout probability of $0.1$ is applied to the sentence representation before passing it to the Softmax layer. Adam~\cite{kingma2014adam} is used for optimization with an initial learning rate of $4e^{-5}$. We use $100$ warmup steps for BERT classifier. We train the model for $8$ epochs and select the best performing model on the dev data. 

All experiments were conducted using a single GPU instance of Nvidia Tesla v100 type. For BART model, we use f16 precision. For all data augmentation models, validation set performance was used to select the best model.

\section{Experimental Setup}
\begin{table*}[ht]
\centering
\begin{tabular}{l|p{14cm}}
\hline 
Data &  Label Names \\ \hline 
SST-2 & Positive, Negative \\ 
TREC  &  Description, Entity, Abbreviation, Human, Location, Numeric \\
SNIPS & PlayMusic, GetWeather, RateBook, SearchScreeningEvent, SearchCreativeWork, AddToPlaylist, BookRestaurant \\
\hline
\end{tabular}
\caption{\label{table:label_names} Label Names used for fine-tuning pre-trained models. Label names are lower-cased for all experiments.}
\end{table*}

\subsection{Baseline Approaches for DA}
In this work, we consider three data augmentation methods as our baselines. 

(1) \textbf{EDA}~\cite{wei2019eda} is a simple word-replacement based augmentation method, which has been shown to improve text classification performance in the low-data regime. 

(2) \textbf{Backtranslation}~\cite{sennrich2015improving} is another widely used text augmentation method~\cite{shleifer2019low,xie2019unsupervised,Edunov2018UnderstandingBA}. For backtranslation, we use a pre-trained EN-DE\footnote{\url{https://dl.fbaipublicfiles.com/fairseq/models/wmt19.en-de.joined-dict.single_model.tar.gz}}, and DE-EN\footnote{\url{https://dl.fbaipublicfiles.com/fairseq/models/wmt19.de-en.joined-dict.single_model.tar.gz}} translation models ~\cite{NG2019facebook}. 

(3) \textbf{CBERT}~\cite{wu2019conditional} language model which, to the best of our knowledge, is the latest model-based augmentation that outperforms other word-replacement based methods.

\subsection{Data Sets}
\label{datasets}

\begin{table}[ht]
\centering
\begin{tabular}{l|r|r|r}
\hline 
& SST-2 & SNIPS & TREC \\ \hline 
Train & 6,228 & 13,084 & 5,406 \\ \hline 
Dev & 692 & 700 & 546 \\ \hline  
Test & 1,821 & 700 & 500\\ \hline 
\end{tabular}
\caption{\label{table:corpus_stat} Data statistics for three corpora, without any sub-sampling. This setup is used to train a classifier for intrinsic evaluation, as described in Section \ref{evaluation}. When simulating low-data regime, we sample 10 or 50 training examples from each category. For testing, we use the full test data.}
\end{table}

We use three text classification data sets. 

(1) \textbf{SST-2}~\cite{socher2013recursive}:  (Stanford Sentiment Treebank) is a dataset for sentiment classification on movie reviews, which are annotated with two labels (Positive and Negative).

(2) \textbf{SNIPS}~\cite{coucke2018snips} dataset contains $7$ intents which are collected from the Snips personal voice assistant. 

(3) \textbf{TREC}~\cite{li2002learning} is a fine-grained question classification dataset sourced from TREC. It contains six question types (whether the question is about person, location, etc.).

For SST-2 and TREC, we use the dataset versions provided by ~\cite{wu2019conditional}\footnote{\url{https://github.com/1024er/cbert_aug}}, and for SNIPS dataset, we use \footnote{\url{https://github.com/MiuLab/SlotGated-SLU/tree/master/data/snips}}. We replace numeric class labels with their text versions. For our experiments, we used the labels provided in Table ~\ref{table:label_names}. Note that pre-trained methods rely on different byte pair encodings that might split labels into multiple tokens. For all experiments, we use the lowercase version of the class labels. 

\subsubsection{Low-resourced data scenario}
Following previous works to simulate the low-data regime setting for text classification \cite{hu2019learning}, we subsample a small training set on each task by randomly selecting an equal number of examples for each class. 

In our preliminary experiments, we evaluated classification performance with various degrees of low-data regime settings, including 10, 50, 100 examples per class. We observed that state-of-the-art classifiers, such as the pre-trained BERT classifier, performs relatively well for these data sets in a moderate low-data regime setting. For example, using 100 training examples per class for SNIPS dataset, BERT classifier achieves 94\% accuracy, without any data augmentation. In order to simulate a realistic low-resourced data setting where we often observe poor performance, we focus on experiments with 10 and 50 examples per class. Note that using a very small dev set leads the model to achieve 100\% accuracy in the first epoch which prevents a fair model selection based on the dev set performance. To avoid this and to have a reliable development set, we select ten validation examples per class.  



\subsection{Evaluation} 
\label{evaluation} 
To evaluate DA, we perform both intrinsic and extrinsic evaluation. For \textbf{extrinsic evaluation}, we add the generated examples into low-data regime training data for each task and evaluate the performance on the full test set. All experiments are repeated 15 times to account for stochasticity. For each experiment, we randomly subsample both training and dev set to simulate a low-data regime.

For \textbf{intrinsic evaluation}, we consider two aspects of the generated text. The first one is semantic fidelity, where we measure how well the generated text retains the meaning and the class information of the input sentence. In order to measure this, we train a classifier on each task by fine-tuning a pre-trained English BERT-base uncased model. Section \ref{intrinsic_evaluation} describes corpus and classifier performance details.

Another aspect we consider is text diversity. To compare different models' ability to generate diverse output, we measured type token ratio \cite{roemmele2017evaluating}. Type token ratio is calculated by dividing the number of unique $n$-grams by the number of all $n$-grams in the generated text. 

 \subsubsection{Classifiers for intrinsic evaluation}
\label{intrinsic_evaluation} 
In this work, we measure semantic fidelity by evaluating how well the generated text retains the meaning and the label information of the input sentence. To measure this, we fine-tune the base classifier described in Section ~\ref{base_classifier}. 

To take full advantage of the labeled data and to make our classifier more accurate, we combine 100\% of training and test partitions of the corresponding dataset, and use the combined data for training. Then, the best classifier is selected based on the performance on the dev partition. Classification accuracy of the best classifier on dev partition for each corpus is provided in Table \ref{table:dev_2}.

\begin{table}[ht] 
\centering
\begin{tabular}{lccc}
\hline 
& SST-2 & SNIPS & TREC \\  \hline 
Dev & 91.91 & 99 & 94.13 \\ \hline 
\end{tabular}
\caption{\label{table:dev_2} Classifier performance on dev set for each corpus. Classifiers are used for intrinsic evaluation. }
\end{table}

\section{Results and Discussion}

\subsection{Generation by Conditioning on Labels}
\label{section:how_to_condition}
As described in Section \ref{approach_bert}, 
we choose BERT as a pre-trained model and explored different ways of conditioning BERT on labels: BERT\textsubscript{prepend},  BERT\textsubscript{expand} and CBERT.  

\begin{table*}[]
\centering
\begin{tabular}{l|c|c|l}
\hline \textbf{Model} & \multicolumn{1}{c|}{\textbf{SST-2}} & \multicolumn{1}{c|}{\textbf{SNIPS}} & \multicolumn{1}{c}{\textbf{TREC}} \\ \hline 
No Aug   & 52.93 (5.01)  &  79.38 (3.20)    &  48.56 (11.53)     \\
EDA  & 53.82 (4.44)     & 85.78 (2.96)     & 52.57 (10.49)    \\
BackTrans.   & 57.45 (5.56)     & 86.45 (2.40)   & 66.16 (8.52)  \\
CBERT   & 57.36 (6.72)       & 85.79 (3.46)    & 64.33 (10.90)  \\
BERT\textsubscript{expand}   & 56.34 (6.48)       & 86.11 (2.70)    & 65.33 (6.05)   \\
BERT\textsubscript{prepend}  & 56.11 (6.33)     & 86.77 (1.61)   & 64.74 (9.61) \\
GPT2\textsubscript{context}  & 55.40 (6.71)    &  86.59 (2.73)   & 54.29 (10.12)   \\
BART\textsubscript{word}     & \textbf{57.97} (6.80)         & 86.78 (2.59) & 63.73 (9.84)   \\
BART\textsubscript{span}     & 57.68 (7.06)     & \textbf{87.24} (1.39) & \textbf{67.30} (6.13)   \\

\hline
\end{tabular}
\caption{\label{low_resource_results} DA extrinsic evaluation in low-data regime. Results are reported as Mean (STD) accuracy on the full test set. Experiments are repeated 15 times on randomly sampled training and dev data. For data augmentation model fine-tuning, we use 10 examples per class for training.}
\end{table*}


Table~\ref{low_resource_results} shows BERT\textsubscript{prepend}, BERT\textsubscript{expand} and CBERT have similar performance on three datasets. Note that BERT is pre-trained on a very huge corpus, but fine-tuning is applied on a limited data. This makes it difficult for the model to learn new, meaningful label representations from scratch as in case the BERT\textsubscript{expand}. While CBERT and BERT\textsubscript{prepend} both converge in less than 8 epochs, BERT\textsubscript{expand} requires more than 100 epochs to converge.  

Further, the class conditioning technique used in CBERT is specific to BERT architecture which relies on modifying BERT's segment embedding and hence cannot be applied to other model architectures.  Since the labels in most of the datasets are well-associated with the meaning of the class (e.g. \textit{SearchCreativeWork}), prepending tokens allows the model to leverage label information for conditional word replacement. Given these insights, we recommend \texttt{prepend} as a preferred technique for pre-trained model based data augmentation.     

 
\subsection{Pre-trained Model Comparison}
\paragraph{Classification Performance} \label{intrinsic_discussion} 
Table~\ref{low_resource_results} shows that seq2seq pre-training based BART outperforms other DA approaches on all data sets. We also observe that back translation (shown as BackTrans. in table) is a very strong baseline as it consistently outperforms several pre-trained data augmentation techniques including CBERT baseline.

\paragraph{Generated Data Fidelity}
As described in Section \ref{intrinsic_evaluation}, we train a classifier for each dataset and use the trained classifier to predict the label of the generated text. 

\begin{table}[h]
\centering
\begin{tabular}{lccc}
\hline \textbf{Model} & \textbf{SST-2} & \textbf{SNIPS} & \textbf{TREC} \\
\hline
EDA                         & 95.00 & 97.14 & 87.22 \\
BackTrans.                    & \textbf{96.66} & 97.14 & \textbf{94.88}  \\
CBERT                       & 96.33 & \textbf{97.90}  & 92.22  \\
BERT\textsubscript{expand}  & 95.00 & 97.04  & 91.44  \\ 
BERT\textsubscript{prepend} & \textbf{96.66} & 97.80 & 94.33  \\
GPT2\textsubscript{context} & 68.33 & 92.47 & 60.77  \\
BART\textsubscript{word}    & 89.33 & 91.03  & 79.33  \\
BART\textsubscript{span}    & 90.66 & 93.04 &  80.22 \\
\hline
\end{tabular}
\caption{\label{intrinsic_fidelity} Semantic fidelity of generated output. We trained a classifier using all labelled data in order to perform accuracy test on the generated data. Higher accuracy score means that the model retains the class label of the input sentence more accurately. For data augmentation model fine-tuning, we use 10 examples per class for training.}
\end{table}

\begin{table}[t]
\small
\centering
\begin{tabular}{l|cc|cc|cc}
\hline \textbf{Model} & \multicolumn{2}{c|}{\textbf{SST-2}} & \multicolumn{2}{c|}{\textbf{SNIPS}} & \multicolumn{2}{c}{\textbf{TREC}} \\ \hline 
\textit{n}-gram & 1 & 3 & 1 & 3 & 1 & 3  \\ 
\hline
EDA                          & 0.66 & 0.99 & 0.49 & \textbf{0.97} & 0.55 & \textbf{0.97}\\ 
BackTrans                    & \textbf{0.68} & 0.99 & \textbf{0.51} & 0.95 & \textbf{0.57} & 0.96 \\ 
CBERT                        & 0.57 & 0.99 & 0.48 & 0.95 & 0.46 & 0.95 \\ 
BERT\textsubscript{expand}   & 0.59 & 0.99 & 0.49 & 0.96 & 0.47 & 0.95 \\ 
BERT\textsubscript{prepend}  & 0.57 & 0.99 & 0.48 & 0.95 & 0.46 & 0.95 \\ 
GPT2\textsubscript{context}  & 0.62 & 0.99 & 0.34 & 0.88 & 0.44 & 0.92 \\ 
BART\textsubscript{word}     & 0.53 & 0.99 & 0.42 & 0.95 & 0.40 & 0.93 \\ 
BART\textsubscript{span}     & 0.55 & 0.99 & 0.41 & 0.91 & 0.39 & 0.89 \\ 
\hline
\end{tabular}

\caption{\label{intrinsic_diversity} Type token ratio for generated text using each model. For data augmentation model fine-tuning, we use 10 examples per class for training.}
\end{table}

Table \ref{intrinsic_fidelity} shows that AE based methods outperform AR models like GPT2, and Seq2seq based model like BART, in terms of semantic fidelity of the generated data. On two datasets, back translation approach outperforms all other methods in terms of fidelity which underlines the effectiveness of the state of the art translation systems in terms of preserving the semantics of the language.   


\paragraph{Generated Data Diversity}
To further analyze the generated data, we explore type token ratio as described in Section \ref{evaluation}. Table \ref{intrinsic_diversity} shows that EDA generates the most diverse tri-grams and back translation approach produces the most diverse unigrams. Since EDA method modifies tokens at random, it leads to more diverse $n$-grams, not necessarily preserving the semantic of the input sentence. Also, unlike AE and Seq2seq methods that rely on word or span replacements, back translation is an open-ended system that often introduces unseen unigrams.

\subsection{Guidelines For Using Different Types Of Pre-trained Models For DA} 
 
\paragraph{AE models}: We found that simply prepending the label to raw sequences provides competitive performance than modifying the model architecture. As expected, more complex AE models such as RoBERTa\textsubscript{prepend} \cite{liu2019roberta} outperforms BERT\textsubscript{prepend} (66.12 vs 64.74 mean acc on TREC).

\paragraph{AR models}: While AR based model such as GPT2 produces very coherent text, it does not preserve the label well. In our experiments, we found that providing a few starting words along with the label as in GPT2\textsubscript{context} is crucial to generate meaningful data. 

\paragraph{Seq2Seq models}: Seq2Seq models provide an opportunity to experiment with various kinds of denoising autoencoder tasks including masking at subword, word or span level, random word insertion or deletion. We observe that word or span masking performs better than other denoising objectives, and should be preferred for DA.

Overall, we found that while AE models are constrained to produce similar length sequences and are good at preserving labels, AR models excel at unconstrained generation but might not retain label information.   
Seq2Seq models lie between AE and AR by providing a good balance between diversity and semantic fidelity. Further, in Seq2Seq models, diversity of the generated data can be controlled by varying the masking ratio.

\subsection{Limitations}
Our paper shows that a pre-trained model can be used for conditional data augmentation by fine-tuning it on the training data where the class labels are prepended to the training examples. Such a unified approach allows utilizing different kinds of pre-trained models for text data augmentation to improve performance in low-resourced tasks. However, as shown in Table ~\ref{50_examples}, improving pre-trained classifier's model performance in rich-resource setting is still challenging. 

Our results also show that a particular pre-trained model based augmentation may do well on one task or dataset, but may not work well for other scenarios. In our experiments, we use the same set of hyperparameters such as masking rate, learning rate and warmup schedule for all three datasets which might not lead to the best performance for all considered tasks. 
While our primary goal in this work is to propose a unified data augmentation technique, we believe that further studies on optimizing performance for a given task or model will be beneficial.

\section{Conclusion And Future Work}
We show that AE, AR, and Seq2Seq pre-trained models can be conditioned on labels by prepending label information and provide an effective way to augment training data. These DA methods can be easily combined with other advances in text content manipulation such as co-training the data generator and the classifier ~\cite{hu2019learning}. Further, the proposed data augmentation techniques can also be combined with latent space augmentation ~\cite{kumar2019closer}. We hope that unifying different DA methods would inspire new approaches for universal NLP data augmentation.

\bibliography{dataaug}
\bibliographystyle{acl_natbib}
\newpage
\appendix

\section{Appendices}
\label{sec:appendix}
\subsection{Classification performance on 50 examples per class}

\begin{table}[h]
\centering \small 
\begin{tabular}{l|c|c|c}
\hline \textbf{Model} & \multicolumn{1}{c|}{\textbf{SST-2}} & \multicolumn{1}{c|}{\textbf{SNIPS}} & \multicolumn{1}{c}{\textbf{TREC}} \\ \hline 
\textit {Num examples} & 50 & 50 & 50  \\ 
\hline
No Aug                       & \textbf{78.60} (2.81)           & 90.98 (2.30)      & 71.65 (11.09)     \\
EDA              & 76.41 (4.90)          & 89.80 (2.99)       & 61.98 (11.52)     \\
BackTrans              & 78.30  (6.30)        & 89.87 (3.35)      & 65.52 (11.82)     \\
CBERT                        & 77.26 (4.56)          & 90.57 (2.11)      &67.77 (13.81)     \\
BERT\textsubscript{expand}   & 78.15 (4.56)          & 89.79 (2.56)      & 68.06 (12.20)     \\
BERT\textsubscript{prepend}  & 77.96 (4.78)          & 90.41 (2.39)      & \textbf{71.88} (9.91)     \\
GPT2\textsubscript{context}  & 74.91 (5.43)          & 88.87 (3.33)      & 51.41 (11.35)     \\
BART\textsubscript{word}     & 76.35 (5.84)          & \textbf{91.40} (1.60) & 71.58 (7.00)     \\
BART\textsubscript{span}     & 77.92 (4.96)           & 90.62 (1.79)      & 67.28 (12.57)    \\

\hline
\end{tabular}
\caption{\label{50_examples} DA extrinsic evaluation in low-data regime. Results are reported as Mean (STD) accuracy on the full test set. Experiments are repeated 15 times on randomly sampled training, dev data. 50 training examples are subsampled randomly.}
\end{table}

Table ~\ref{50_examples} shows the classification performance of different models when we select 50 examples per class. Overall, we find that data augmentation does not improve classification performance, and in many cases, it even hurts the accuracy.   

\end{document}